# Soil Image Segmentation Based on Mask R-CNN

Authors Name/s per 1st Affiliation (Author)* Yida Chen[1,2,a]
[1]College of Computer and Information Science, Chongqing Normal University, Chongqing, China
[2]Chongqing Center of Engineering Technology Research on Digital Agricultural Service, Chongqing, China
* Corresponding author: [a]yida_myth@stu.cqnu.edu.cn

Authors Name/s per 2nd Affiliation (Author) Kang Liu[3,4,b]
[3]College of Computer and Information Science, Chongqing Normal University, Chongqing, China
[4]Chongqing Center of Engineering Technology Research on Digital Agricultural Service, Chongqing, China
[b]e-mail: 1914900130@qq.com

Authors Name/s per 3rd Affiliation (Author) Yi Xin[5,c]
[5]Agriculture and Rural Committee of Nanan District, Chongqing, China
[c]e-mail: 442850911@qq.com

Authors Name/s per 4th Affiliation (Author) Xinru Zhao[6,d]
[6]Agriculture and Rural Committee of Banan District, Chongqing, China
[d]e-mail: 871354656@163.com

*Abstract*—The complex background in the soil image collected in the field natural environment will affect the subsequent soil image recognition based on machine vision. Segmenting the soil center area from the soil image can eliminate the influence of the complex background, which is an important preprocessing work for subsequent soil image recognition. For the first time, the deep learning method was applied to soil image segmentation, and the Mask R-CNN model was selected to complete the positioning and segmentation of soil images. Construct a soil image dataset based on the collected soil images, use the EISeg annotation tool to mark the soil area as soil, and save the annotation information; train the Mask R-CNN soil image instance segmentation model. The trained model can obtain accurate segmentation results for soil images, and can show good performance on soil images collected in different environments; the trained instance segmentation model has a loss value of 0.1999 in the training set, and the mAP of the validation set segmentation (IoU=0.5) is 0.8804, and it takes only 0.06s to complete image segmentation based on GPU acceleration, which can meet the real-time segmentation and detection of soil images in the field under natural conditions. You can get our code in the Conclusions. The homepage is https://github.com/YidaMyth.

I. INTRODUCTION

Soil is an indispensable part of agriculture. Correct identification of soil types can guide crop production and planting and maximize economic benefits. It is very difficult to directly recognize purple soil images collected in the field under natural conditions based on machine vision technology[1]. Therefore, separating the soil area from the complex background in the field is the primary work for identifying soil types.

At present, there are few researches on purple soil image segmentation, and they all stay in the traditional image segmentation method. Cheng et al.[2] proved that purple soil has good separation characteristics in the H domain through prior knowledge, determined the H threshold based on the normal distribution, and segmented the soil and the background area; Zeng et al.[3] optimized the confidence probability P and the H domain segmentation threshold, and proposed A segmentation algorithm based on Chebyshev's inequality to further improve the segmentation accuracy of purple soil images; Wu et al[4]. improved the SLIC algorithm to roughly segment purple soil, then redefined the merge threshold to merge purple soil superpixels, and finally proposed a filling algorithm to eliminate soil Cavities inside the area can effectively extract the complete image of the purple soil area; Zeng et al.[5] and others realized the color image segmentation algorithm of purple soil in the field based on adaptive density peak clustering. The density peak clustering algorithm adaptively segmented the soil image, and finally designed a post-processing algorithm to obtain the complete soil area to further improve the segmentation accuracy of the soil image; Yang et al.[6] used the RGB color space to construct a segmented separation measure, and optimized the local area based on the density peak clustering idea. Segmentation threshold, finally eliminate connected areas and fill holes to obtain soil image segmentation results; Zeng et al.[7] improved the FCM algorithm and proposed a SWFCM algorithm to accurately segment the boundaries of soil and non-soil areas and obtain accurate soil image segmentation results .

Subsequent research on soil type identification requires image segmentation to eliminate the influence of complex backgrounds. However, when image segmentation is based on traditional algorithms, the robustness of the algorithm is limited in different complex environments, and the segmentation effect is not good. Therefore, this paper applies deep learning methods to soil image segmentation research for the first time.

II. SOIL IMAGE COLLECTION AND LABELING

*A. Soil Image Data Acquisition*

According to Chongqing Soil Classification and Code DB50/T796-2017, 34 soil species of 3 major soil genera (dark purple mud, red brown purple mud, gray brown purple mud) distributed in Bishan District were studied. In the field, use a soil spade to shovel out the purple soil of 0-20cm in the plow layer, and take pictures of the natural fracture section of the purple soil (core soil), so that the purple soil area is located in the center of the image.

During the collection process, the soil images are labeled under the identification of experts for subsequent identification research. The research team of our laboratory collected soil images under different lighting conditions (including sunny, cloudy, and cloudy) in August, September, October, and November 2020. When collecting images, adjust the camera parameters, use the mobile phone rack to fix the device, stipulate that the soil center is placed in the camera, so that the soil occupies more than 50% of the entire image, and finally control the remote Bluetooth device to take soil images from multiple angles.

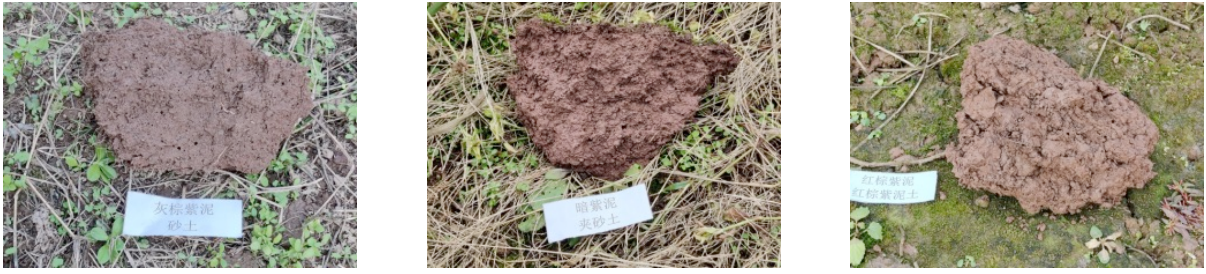

Figure 1. Original of soil image

### B. Soil Segmentation Dataset

A total of 111 soil images were collected in August and September. 78 soil images were randomly selected at 7:3 to form the training set, and 33 soil images were used to form the verification set. After the model training was completed, the soil images taken in October and November were randomly selected for the final test.

### C. Data Annotation

Use EISeg of Baidu Feijiang to label the soil image, record the soil area label as soil, and save the label data as the corresponding COCO2017 dataset format; label the training set and verification set images respectively, and then save the label information as instances_train2017.json, and instances_val2017.json into the annotations folder; the corresponding training set and validation set images are moved to the train2017 and val2017 folders respectively.

## III. MATERIALS AND METHODS

### A. Mask R-CNN Instance Segmentation Model

In order to eliminate the influence of complex environment in subsequent soil type identification, soil image segmentation is a necessary step before identification. At this stage, the soil image segmentation methods still stay on the traditional segmentation algorithm, either based on threshold segmentation; or construct a separable measure. Such a method has certain limitations and can only achieve good results on targeted data sets. However, it is difficult to obtain ideal results under complex field natural conditions in an unknown environment, and the processing speed of high-resolution images is slow and the segmentation accuracy is limited; it is difficult to process the original soil image based on traditional image segmentation algorithms, and it takes a lot of time. Therefore, this paper applies the deep learning method to the research of soil image segmentation for the first time.

Mask R-CNN[8] is very suitable for related research on soil images. Based on this model, the soil images collected under natural conditions in the field can be segmented; the segmentation results will be used in subsequent recognition research.

Therefore, to train the Mask R-CNN network model to segment the soil image, the overall process of the network is shown in the Fig. 2:

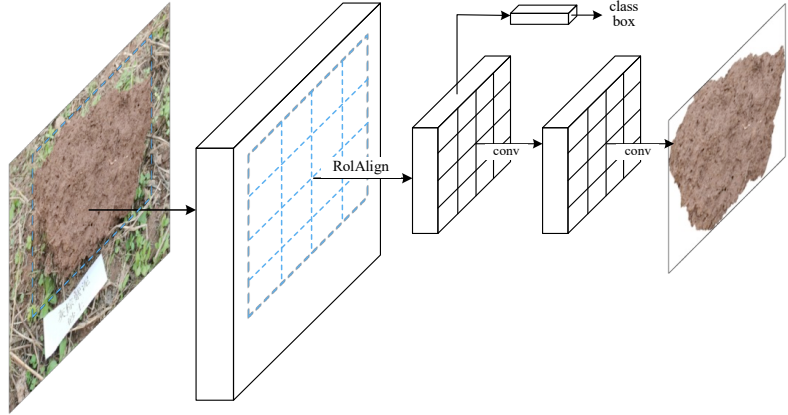

Figure 2. Model structure of Mask R-CNN

### B. Detailed Structure of the Model

Mask R-CNN is based on Faster R-CNN[9], which adds a parallel segmentation Mask branch for instance segmentation, and replaces the ROI pooling layer in Faster R-CNN with a more accurate ROI Align layer. Through the convolutional neural network Feature extraction is performed on the input image, and then the ROI is scaled to a fixed feature map size through the RoI pooling layer, but two quantizations will lose part of the image information, which has little impact on the target detection task, but cannot achieve accurate segmentation in instance segmentation , so a more efficient RoI Align layer is used in Mask R-CNN, which effectively avoids the loss caused by two quantizations through ingenious difference operations, and effectively improves the performance of the model; the Mask branch is used for segmentation, and the Faster R-CNN distinguishes the objects of this category for segmentation and obtains the segmentation results.

The loss of Mask R-CNN is to add the loss of Mask branch on the basis of Faster R-CNN. The loss function is:

$$Loss = L_{rpn} + L_{faster\_rcnn} + L_{mask} \qquad (1)$$

In the first part, RPN is the core of Faster R-CNN. For the first time, CNN is used to perform Region Proposal to obtain target candidate boxes. RPN will preset k anchors, and then obtain 2k categories and 4k regression parameters through convolution for the loss calculation of the boundary regression box.

In the second part, the loss of Faster R-CNN consists of two parts, the first part is the classification loss, and the second part is the bounding box regression loss.

The third part, the loss on the Mask branch is the binary cross entropy loss (Binary Cross Entropy) between the predicted mask and the real label.

The specific architecture of Mask R-CNN is shown in the Fig. 3:

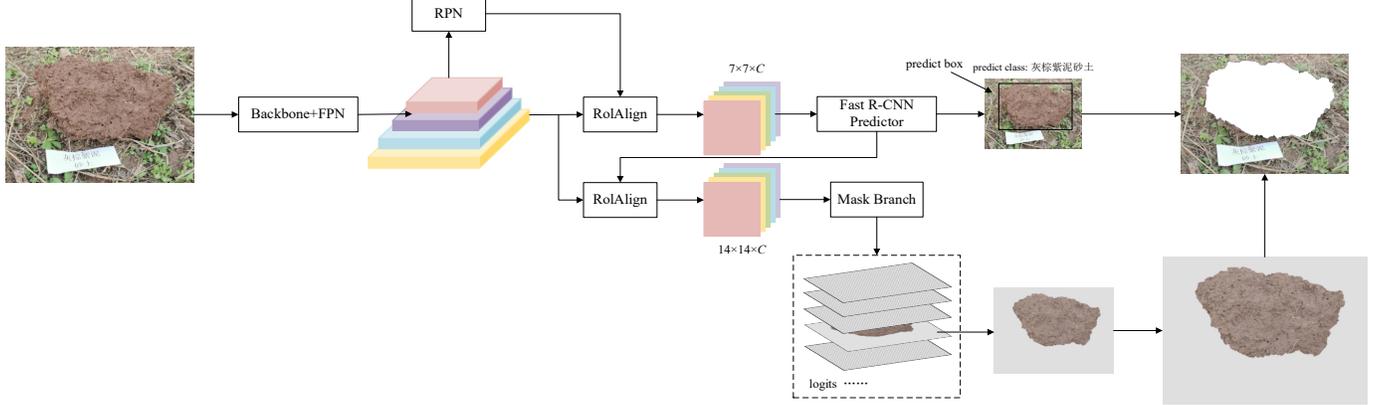

Figure 3. Detailed structure of soil image segmentation model

For the input image, the feature extraction is performed through the backbone network FPN first, and then the candidate frame is generated by the RPN and then scaled to a fixed size by ROI Align. Classification and bounding box regression are performed through the Faster R-CNN branch, and the corresponding category results are predicted in the Mask branch. The mask and the real label and the corresponding area are cropped and then the loss is calculated. After the model training is completed, the Faster R-CNN branch performs classification and bounding box regression, and uses the Mask branch to calculate the instance segmentation results of the corresponding target.

## IV RESULTS & DISCUSSION

### A. Experimental Environment and Parameters

1) Experimental environment: server Ubuntu20.04 LTS, 64-bit, processor Intel® Xeon(R) Silver 4114 CPU @ 2.20GHz, memory 64GB, graphics card TITAN V memory 12G; PyCharm Professional Edition 2021.1, Python3.8.5, torch1.10, torchvision0.11.1, numpy1.19.5, matplotlib3.3.3. (Model training and testing both use GPU accelerated computing)

2) Experimental parameters: Epoch is 25, learning rate learning_rate=0.004, optimizer is SGD, momentum momentum=0.9, weight decay weight_decay=0.0001, learning rate strategy uses linear decay: shrink by 0.1 times at the 10th and 20th epoch , batch_size=3, use mixed precision training when training the model, and use transfer learning to load the COCO pre-trained backbone model resnet50; the experimental labeling tool is: EISeg.

3) Experimental objects: Use part of the data collected in 2020 as training samples, a total of 78 pieces; use the newly collected data in 2021 as test samples; use training and test samples with different distributions to test the real performance of the model and verify the model robustness.

### B. Experimental Results and Analysis

During the model training process, the change of the model learning rate and the training loss during the training process are recorded; at the same time, the average mAP value on the test set is counted; mAP is Mean Average Precision, which is used to measure the performance of the Segmentation model, and the value is 0 Between 1, the larger the value, the better the performance. The experimental results are shown in the Fig. 4:

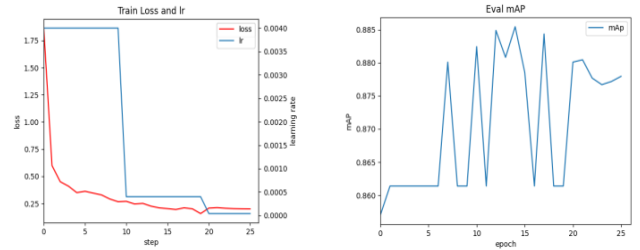

Figure 4. The training process of the model and eval mAP

The soil image is an irregular object, and the edge is not smooth; the background of the soil image taken in the field is extremely challenging, and the training curve will fluctuate normally during the model training phase.

After saving the model, it is used for data testing of soil images. The test results are shown in Fig. 5:

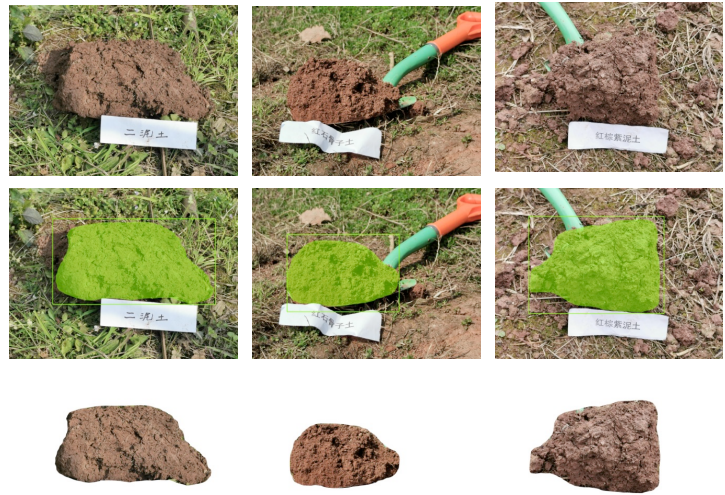

Figure 5.  Test results of soil image segmentation collected in the field

The trained model only needs 0.06s to run on the GPU; the model can achieve good segmentation results in various wild natural scenes. Model testing on soil images collected in different periods has strong robustness.

$$\begin{cases} result = Original \otimes mask \\ mask\_rcnn = Rectangle \cap result \end{cases} \quad (2)$$

Original is the original image of the soil, mask is the mask output by the training model, and a binary label matrix (the current category pixel is marked as 1, and the remaining pixels are marked as 0); it is obtained by Hadamard product on the original image to obtain the segmentation result (and the background is reversed, the pixel value is set to 255, white); through the intersection of the prediction frame Rectangle obtained by the model detection and the segmentation result, the minimum circumscribed rectangle of the soil image is obtained, which is used as the output of the soil image segmentation and can be used for subsequent soil types Identify studies.

## V. CONCLUSIONS

Based on deep learning methods, it can easily solve areas that cannot be effectively solved by traditional methods. The trained Mask R-CNN model can accurately segment soil images and do a good job of preprocessing for subsequent soil image recognition research. The transition from traditional methods to deep learning methods is a necessary journey for the development of downstream vision tasks. Readers can get the code, data annotation and model training methods in our blog: https://yidamyth.blog.csdn.net/article/details/124851003.

## ACKNOWLEDGMENT

Special thanks to the following funding projects for their support for this research:

- Key Science and Technology Research Program (No. KJZD-K2019005005) of Chongqing, China.
- Chongqing University Innovation Research Group funding (No. CXQT20015) of Chongqing Municipal Education Commission, China.
- Chongqing Normal University Postgraduate Scientific Research Innovation Project (YKC22016), China.

## REFERENCES

[1] Srivastava P, Shukla A, Bansal A. (2021) A comprehensive review on soil classification using deep learning and computer vision techniques. J. Multimedia Tools and Applications., 80(10): 14887-14914.

[2] Rong Cheng, Shaohua Zeng, Yutong Luo. (2019) The Color Image Segmentation of Purple Soil with Its H Threshold. J. Journal of Chongqing Normal University(Natural Science)., 36(02): 86-95.

[3] Shaohua Zeng, Yutong Luo, Shengming Yang. (2019) Color Image Segmentation of Purple Soil Based on Chebyshev Inequality. J. Journal of Southwest University (Natural Science Edition)., 41(08): 141-150.

[4] Yalan Wu, Shaohua Zeng, Zhuohua Zeng. (2019) Color Image Segmentation of Purple Soil Based on Improved SLIC. J. Journal of Chongqing Normal University(Natural Science)., 36(05): 106-116.

[5] Shaohua Zeng, Wenmi Tang, Linqing Zhan. (2019) Color image segmentation of field purple soil based on adaptive density peaks clustering. J. Transactions of the Chinese Society of Agricultural Engineering., 35(19): 200-208.

[6] Shengming Yang, Shaohua Zeng, Jingkun Zhao. (2020) Segmentation and Extraction Methods of Purple Soil Image. J. Journal of Chongqing Normal University(Natural Science)., 37(04): 114-123+148.

[7] Zeng S, Wu Y, Wang S. (2021) Adaptive scale weighted fuzzy C-Means clustering for the segmentation of purple soil color image. J. Journal of Intelligent & Fuzzy Systems., 40(6): 11201-11215.

[8] He K, Gkioxari G, Dollár P. (2017) Mask r-cnn. In: Proceedings of the IEEE international conference on computer vision. Honolulu, HI, USA. pp. 2961-2969.

[9] Girshick R. (2015) Fast r-cnn. In: Proceedings of the IEEE international conference on computer vision. Santiago, Chile. pp. 1440-1448.